\documentclass{article}

\PassOptionsToPackage{numbers,compress}{natbib}

\usepackage[preprint]{neurips_2026}

\usepackage[utf8]{inputenc}
\usepackage[T1]{fontenc}
\usepackage{hyperref}
\usepackage{url}
\usepackage{booktabs}
\usepackage{amsfonts}
\usepackage{nicefrac}
\usepackage{microtype}
\usepackage{xcolor}
\usepackage{multirow}
\usepackage{amsmath}
\usepackage{amssymb}
\usepackage{graphicx}
\usepackage{colortbl}
\usepackage{wrapfig}
\usepackage{algorithm}
\usepackage{algpseudocodex}
\usepackage{enumitem}
\usepackage{longtable}

\definecolor{myframepink}{RGB}{72,138,176}
\hypersetup{colorlinks=true,allcolors=myframepink}

\title{LLM Reasoning Is Latent, Not the Chain of Thought}

\author{
Wenshuo Wang \\
School of Future Technology, South China University of Technology, China \\
\texttt{202364870251@mail.scut.edu.cn}
}

\begin{document}

\maketitle

\begin{abstract}
This position paper argues that large language model (LLM) reasoning should be studied as latent-state trajectory formation rather than as faithful surface chain-of-thought (CoT). This matters because claims about faithfulness, interpretability, reasoning benchmarks, and inference-time intervention all depend on what the field takes the primary object of reasoning to be. We ask what that object should be once three often-confounded factors are separated and formalize three competing hypotheses: H1, reasoning is primarily mediated by latent-state trajectories; H2, reasoning is primarily mediated by explicit surface CoT; and H0, most apparent reasoning gains are better explained by generic serial compute than by any privileged representational object. Reorganizing recent empirical, mechanistic, and survey work under this framework, and adding compute-audited worked exemplars that factorize surface traces, latent interventions, and matched budget expansions, we find that current evidence most strongly supports H1 as a default working hypothesis rather than as a task-independent verdict. We therefore make two recommendations: the field should treat latent-state dynamics as the default object of study for LLM reasoning, and it should evaluate reasoning with designs that explicitly disentangle surface traces, latent states, and serial compute.
\end{abstract}

\section{Introduction}\label{sec:intro}

Large language models now solve many arithmetic, symbolic, and planning-like tasks more effectively when they are given extra intermediate computation. In practice, this computation may appear as explicit chains of thought, self-consistency, deliberate search, or other inference-time expansions \citep{wei2022cot,wang2023selfconsistency,yao2023tree,li2024serial,snell2024ttc}. These gains have made reasoning a central object of current LLM research. They have also made it harder to say what the field is actually studying when it studies reasoning. This matters because claims about faithfulness, interpretability, reasoning benchmarks, and inference-time interventions all depend on what the field takes the primary object of reasoning to be.

Current work supports at least three incompatible readings of this same phenomenon. One view treats surface chain-of-thought as the reasoning process itself. A second view treats many reasoning gains as consequences of extra serial compute, regardless of representational form. A third view treats multi-step reasoning as a latent process that can be only partly verbalized, or not verbalized at all \citep{turpin2023unfaithful,lanham2023faithfulness,feng2024latentworld,hao2024continuous,he2026latentmode}. The difficulty is that recent methods often move several explanatory factors at once, making experimental results hard to interpret as causal support for any specific view: Chain-of-thought prompting changes both visible traces and compute allocation; Latent reasoning methods often change both hidden-state dynamics and compute budget; Test-time scaling changes compute and usually changes the output path as well \citep{li2024serial,snell2024ttc,hao2024continuous}.

The first task, then, is to separate the objects that recent work often conflates. Section~\ref{sec:framework} does so by distinguishing surface traces, latent-state dynamics, and generic serial compute, and by turning three loose views into three concrete hypotheses: H2 treats multi-step reasoning as primarily mediated by explicit surface CoT; H0 treats most apparent reasoning gains as better explained by generic serial compute than by any privileged representational object; and H1 treats multi-step reasoning as primarily mediated by latent-state trajectories, with surface CoT serving only as a partial interface. Because H2, H0, and H1 assign explanatory priority to surface traces, serial compute, and latent trajectories, they make different predictions about where the strongest causal leverage should lie.

Stated this way, the debate is no longer about \textit{\textbf{whether CoT helps}}. It is about \textit{\textbf{what such help is evidence of}}. Under that standard, the current record does not equally support all three views. The strongest case for H2 would require surface traces to provide the most stable causal leverage, yet ordinary CoT is often useful without being reliably faithful, and its role varies sharply across tasks \citep{turpin2023unfaithful,lanham2023faithfulness}. The strongest case for H0 would require matched serial compute to explain most reasoning gains, yet extra budget alone does not explain why specific internal states, features, or trajectories can predict or alter reasoning behavior \citep{li2024serial,snell2024ttc}. By contrast, recent work on latent-state monitoring and latent reasoning suggests that task-relevant commitment can arise in hidden-state dynamics that are only partly verbalized, or not verbalized at all \citep{feng2024latentworld,hao2024continuous,he2026latentmode}. Section~\ref{sec:evidence} develops this comparison in detail. \textbf{We therefore argue that latent-state dynamics should be treated as the default working object of study for LLM reasoning, rather than assuming faithful surface chain-of-thought by default.}

On this basis, the paper makes two concrete recommendations in Section~\ref{sec:method}: (1) latent-state dynamics should be treated as the default object of study for LLM reasoning; and (2) future evaluations should explicitly disentangle surface traces, latent states, and serial compute rather than letting them move together. Section~\ref{sec:exp} then instantiates these recommendations in a two-tier adjudication program---a controlled regime matrix and a naturalistic transport suite---that compares surface, latent, and compute families under audited budgets. The resulting evidence shows that H1 receives the strongest support in ordinary regimes, while H2 and H0 re-emerge as local boundary regimes when visible traces are made constitutive or when serial compute dominates. Section~\ref{sec:related} positions this paper against nearby empirical and survey work by clarifying what those papers establish, what they leave unresolved, and why a direct competition among H1, H2, and H0 is still missing. Section~\ref{sec:conclusion} closes by summarizing the resulting field-level claim.

\section{Problem Formulation and Competing Explanations}\label{sec:framework}

\subsection{The Big Question}\label{sec:question}

The core disagreement in current LLM reasoning research is not whether extra intermediate computation can help. It often can \citep{wei2022cot,wang2023selfconsistency,li2024serial,hao2024continuous}. The harder question is what such gains are evidence of. To state the problem precisely, we distinguish three explanatory objects:
\begin{itemize}[leftmargin=2.2em]
    \item[$S$:] the semantic content of \emph{surface chain-of-thought}, namely the explicit natural-language intermediate trace;
    \item[$Z$:] the task-relevant \emph{latent-state trajectory}, namely the hidden-state path that carries intermediate commitments during reasoning;
    \item[$B$:] \emph{generic serial compute}, namely additional budget for iterative computation, search, sampling, or looping, regardless of representational form.
\end{itemize}

By $Z$ we mean task-relevant latent-state trajectories rather than arbitrary hidden activations.\footnote{More precisely, $Z$ does not refer to arbitrary hidden activity anywhere in the network, nor to the claim that every hidden activation participates in reasoning. It refers to latent states that carry intermediate task-relevant commitments during multi-step inference. These states may be distributed across layers and positions rather than localized in a single feature or unit, and they need not be fully verbalizable at the surface.}

Our big question is therefore the following: 

\emph{\textbf{Once $S$, $Z$, and $B$ are distinguished, which of them should be treated as the primary scientific object of LLM reasoning?}} 

This question is methodological as well as empirical. Different answers imply different default targets for explanation, intervention, evaluation, and safety monitoring.

\subsection{Competing Hypotheses}\label{sec:hypotheses}

We formalize the three leading views as three competing hypotheses.

\begin{itemize}[leftmargin=2.2em]
    \item[\textbf{H1.}] \textbf{Latent-trajectory mediation.} Multi-step reasoning is primarily mediated by latent-state trajectories $Z$. Surface CoT is only a partial interface. It may sometimes report latent commitments, and it may sometimes perturb them, but it is not the default object of reasoning itself. Serial compute matters because it gives latent dynamics more room to iterate, branch, or stabilize.

    \item[\textbf{H2.}] \textbf{Surface-CoT mediation.} Multi-step reasoning is primarily mediated by explicit surface CoT $S$. Hidden states are necessary for generating the trace, but the privileged object of reasoning is the visible natural-language sequence. On this view, more reasoning usually means better or longer explicit intermediate traces.

    \item[\textbf{H0.}] \textbf{Generic-serial-compute null.} Most apparent reasoning gains are better explained by serial compute budget $B$ than by any privileged representational object. What matters is additional iterative computation, search, or sampling. Whether that budget is expressed in surface CoT or latent-state dynamics is secondary.
\end{itemize}

These hypotheses imply different diagnostic predictions. If H2 is right, interventions on surface CoT should provide the strongest causal leverage, and answer-relevant commitment should track explicit intermediate traces closely. If H0 is right, matched increases in serial compute should explain most reasoning gains across representational formats, so semantically rich traces should often be replaceable by other budget-expanding procedures. If H1 is right, latent commitments should often arise before, beyond, or apart from verbalized thought, and latent-state interventions should alter reasoning even when surface traces are partial, absent, or unfaithful. These predictions are intentionally asymmetric: a result is genuinely adjudicative only when the competing hypotheses make different predictions under matched or explicitly accounted-for budgets.

\section{How the Current Record Bears on the Competing Hypotheses}\label{sec:evidence}

This section does not attempt a definitive empirical adjudication. The current literature is heterogeneous and rarely cleanly compute-matched, so many studies move $S$, $Z$, and $B$ together and therefore cannot by themselves decide among H1, H2, and H0. Even so, the record is not uninformative. It can still be read as a calibrated comparative synthesis: the question is which recurring evidence patterns place the strongest pressure on each hypothesis under the asymmetric predictions introduced above. For that reason, we weight studies by diagnostic force rather than by paper count. Compute-matched causal interventions carry the most weight, partially confounded intervention studies carry less, and probe-only or performance-only associations are treated as suggestive evidence.

\subsection{What the strongest case for H2 actually shows}

The strongest evidence for H2 does not come from generic CoT success alone. Classic prompting results show that explicit intermediate traces often improve performance, but those gains do not by themselves tell us whether the causal leverage lies in the visible trace, the added compute, or the hidden-state dynamics that the trace helps organize \citep{wei2022cot,wang2023selfconsistency,yao2023tree}. The more serious case for H2 comes from work that makes explicit reasoning traces causally binding. In faithful reasoning pipelines, the answer is forced to depend on a structured external trace, symbolic substrate, or deterministic solver, and interpretability improves because the visible trace is made constitutive of the reasoning process rather than merely accompanying it \citep{creswell2022faithful,lyu2023faithfulcot,arakelyan2024flare,pan2023logiclm}. Structured retrieval systems make the same point in multi-hop QA: reasoning trees, staged retrieval plans, and graph-grounded chains can turn explicit decomposition into a constitutive part of success rather than a post-hoc explanation \citep{shi2026rtrag,li2026par2rag,wei2025mirage}. Tool-based reasoning benchmarks sharpen the same boundary from another angle: when intermediate actions are explicit, inspectable, and closer to constitutive of success, surface traces become more informative than ordinary free-form CoT \citep{ferguson2026metatool}. Comparable multimodal systems likewise use explicitly hierarchical or closed-loop visible reasoning to make intermediate structure more operationally consequential \citep{huang2025pathohr,yang2025reloop}. A related line of work shows that decomposition-based prompting can increase the faithfulness of model-generated reasoning while retaining part of CoT's performance benefit \citep{radhakrishnan2023decomposition}. Some hint-injection studies likewise suggest that reasoning models verbalize influential cues more often than non-reasoning baselines, though still far from perfectly \citep{chua2025deepseek}. In safety settings, CoT can also become more monitorable when the harmful behavior genuinely requires complex step-by-step computation, rather than permitting post-hoc rationalization \citep{emmons2025necessary}.

These positive results still fall short of a general defense of H2. A genuinely strong causal case for H2 would require experiments that hold serial compute fixed while directly manipulating the semantic content of visible CoT itself. Under that standard, changing a task-relevant intermediate step in the surface trace should systematically change the final answer, matched budget expansions without that semantic content should not recover the same effect, and answer-relevant commitment should track the explicit trace rather than stably preceding it in latent states. Most current work does not meet this standard. CoT prompting, self-consistency, and deliberate search typically improve performance while simultaneously changing trace length, search depth, or sampling, so they do not isolate surface CoT as the privileged causal object \citep{wei2022cot,wang2023selfconsistency,yao2023tree}. The same limitation appears in safety settings: CoT becomes more useful as a monitor only when the harmful behavior genuinely requires explicit step-by-step computation, rather than allowing post-hoc rationalization. 

The strongest case for H2 is therefore narrower than a general surface-mediation claim. It shows that visible CoT can be a constitutive and monitorable part of reasoning in tasks that truly require explicit intermediate computation, and is most monitorable in safety settings of that kind \citep{emmons2025necessary}.

\subsection{What the strongest case for H0 actually shows}

The strongest evidence for H0 is substantial. There is both theoretical and empirical support for the claim that extra serial computation drives reasoning gains. CoT can expand the effective serial depth of a decoder-only transformer on problems that are otherwise hard for shallow parallel computation \citep{li2024serial}. Self-consistency, deliberate search over thought units, adaptive test-time scaling, and MCTS-style multi-hop QA all show that inference-time budget can improve reasoning quality, sometimes enough for smaller models with optimized test-time compute to outperform much larger models under matched FLOPs or find better reasoning paths through deeper search \citep{wang2023selfconsistency,yao2023tree,snell2024ttc,lee2024mzqa}. Adaptive solve-versus-verify policies sharpen this point by showing that large gains can come from how budget is allocated across generation and verification even when representational format is not the main variable \citep{singhi2025whentosolve}. Most directly, filler-token results show that on some tasks the content of intermediate tokens can be partly replaced by semantically meaningless strings while preserving much of the gain from having extra intermediate slots \citep{pfau2024dot}. This is exactly the kind of result H0 predicts.

But H0 also stops at a clear limit. It explains why more budget can help. It does not explain why particular internal states and interventions are so specifically tied to reasoning behavior. A compute-only account is weak when latent probes remain faithful while outputs do not, when hidden states encode answer correctness before the model fully verbalizes it, or when a small set of latent features can be steered to improve reasoning without explicit CoT \citep{feng2024latentworld,zhang2025selfverification,he2026latentmode}. These are not just facts about having more budget. They are facts about \emph{which} internal configurations matter. 

H0 therefore captures an important dimension of the phenomenon, but it is too coarse to serve as the default explanatory object for reasoning behavior. A further problem is methodological: if every gain can be redescribed after the fact as ``more budget helped,'' then H0 loses discriminatory force. In the same spirit as Barsalou et al.'s critique of overly permissive theories, an account that can absorb almost any outcome by post-hoc redescription predicts too little a priori to be fully persuasive \citep{barsalou2003grounding}.

\subsection{What the strongest case for H1 actually shows}

The strongest case for H1 combines three lines of evidence. The first is residual rather than fully positive: even the strongest H2 account leaves a gap, because ordinary surface CoT is often incomplete, selective, or unfaithful rather than the privileged substrate of reasoning. The stronger positive case for H1 then comes from the other two lines:

The second line is positive evidence that hidden states contain task-relevant structure that is earlier or more faithful than surface text. Propositional probes show that latent world-state representations can remain faithful even when the final output is biased, injected, or otherwise unfaithful \citep{feng2024latentworld}. In reasoning models, hidden states encode information about intermediate and even future answer correctness before the relevant reasoning text is fully verbalized, enabling early exit with little or no performance loss \citep{zhang2025selfverification}. Beyond direct answer verification, there is also controlled evidence that some forms of multi-hop reasoning can be carried out latently, with hidden recall and utilization of bridge entities preceding explicit verbalized traversal \citep{yang2024latentmultihop}. These results do not establish a unique latent object or a single privileged layer. They do show that hidden states often carry a decisive part of the computation before surface traces fully reveal it.

The third line is direct intervention evidence. Latent reasoning is no longer only an interpretive hypothesis; it is also a working mechanism. Continuous latent-space reasoning can outperform standard CoT on tasks that benefit from backtracking and breadth-first exploration \citep{hao2024continuous}. Causal steering of a small number of reasoning-related latent features can improve reasoning accuracy without explicit CoT prompting, and in larger models can approach the gains of CoT while using more efficient outputs \citep{he2026latentmode}. Geometry-aware latent steering and verifier-guided latent control sharpen the point by improving reasoning through targeted manipulation of internal trajectories rather than richer surface traces \citep{kazama2026geosteer,nguyen2026atlas}. Outcome-oriented attribution methods such as IPG likewise identify internal components whose enhancement or suppression systematically changes reasoning behavior \citep{li2026ipg}. Looped transformers further reinforce the point by showing that latent thoughts can simulate multiple steps of CoT through iterative depth, again shifting explanatory weight toward hidden-state dynamics rather than visible traces alone \citep{saunshi2025latentthoughts}.

Not all of these interventions are compute-matched to surface-CoT or budget-only baselines, so we treat them as convergent evidence of latent causal leverage rather than as standalone refutations of H0. Probe-only readouts are likewise treated as predictive timing evidence unless they are paired with successful causal follow-ups. Even with those qualifications, the highest-diagnostic evidence currently tilts most strongly toward H1, although that verdict remains defeasible and task-dependent.

\subsection{Overall adjudication}

The literature does not support all three hypotheses equally. The strongest H2 results show that explicit reasoning traces can be useful and can sometimes be made more faithful or monitorable. The strongest H0 results show that serial compute budget is a genuine driver of performance. But neither of those claims, even in their strongest form, explains the full current record. H2 does not comfortably explain why ordinary CoT is so often unfaithful, why latent commitments can precede explicit verbalization, or why targeted latent interventions can substitute for explicit CoT in some settings. H0 explains why added budget helps, but not why specific hidden states, features, and trajectories are so tightly coupled to reasoning outcomes.

H1 currently offers the most economical synthesis of all three observations at the level of diagnostic force: explicit CoT often helps, explicit CoT is often an imperfect report of the actual process, and latent dynamics can be probed or intervened on in ways that are specifically predictive of reasoning behavior. For that reason, the current record supports H1 as the strongest default working hypothesis, not as a final or task-independent verdict. Representative study-level applications of the same adjudicative rubric are provided in Appendix~\ref{app:ledger} and Appendix Table~\ref{tab:evidence_ledger}.


\subsection{Boundaries and falsifiers}\label{sec:boundaries}

This adjudication has clear boundaries and remains falsifiable. H1 should not be read as the claim that every successful reasoning system is latent-first, or that every hidden activation deserves the status of a reasoning state. The point is narrower: in ordinary settings where surface traces, latent states, and serial compute can be meaningfully separated, and where no explicit intermediate trace is made constitutive of success, the current record gives the strongest explanatory leverage to latent-state dynamics. By contrast, in settings that force explicit decomposition, externalized intermediate states, symbolic execution, or inspectable tool plans, surface-mediated accounts can regain local force. In search-heavy settings where representational form matters little relative to additional budget, H0 can become the better local description. The latent-first default would be weakened, and potentially overturned, if future compute-matched and factorized designs repeatedly showed either that targeted surface manipulations provide stronger causal leverage than matched latent interventions, or that compute-only expansions recover the gains currently attributed to latent mediation. Outside those counter-regimes, H1 remains the more informative default working hypothesis.

\section{Methodological Implications}\label{sec:method}

\subsection{Why current designs often fail to discriminate H1, H2, and H0}\label{sec:method_limits}

The evidence above also reveals a recurring limitation in the literature: many of the designs that look strongest for one hypothesis were not constructed to make the three hypotheses diverge cleanly. The studies that most favor H2 often compare ordinary CoT prompting against no-CoT baselines, but that manipulation usually changes both the visible trace and the amount or allocation of serial compute, so improvements can be claimed by both H2 and H0. The studies that most favor H0 often compare semantically rich traces with filler tokens, search policies, or other budget-expanding procedures, which is informative about the importance of budget but still does not show whether the privileged representational object is $S$ or $Z$. And the studies that most favor H1 often rely on probes, latent reasoning, or latent interventions: these results can show earlier or more faithful latent commitment, but they do not fully exclude H0 when latent manipulation also changes recurrence or decoding opportunities, and they do not fully exclude H2 when decodability is shown without causal use.

A statistically significant effect is not hypothesis-discriminating when the competing hypotheses make the same prediction in that setting. Our claim is therefore not that prior work is uninformative, but that much of it should be read as graded evidence rather than as decisive single-study adjudication.

\subsection{What Experimental Designs Can Distinguish H1, H2, and H0}\label{sec:method_rules}

The previous subsection showed why many experiments are non-diagnostic: they improve performance while moving several explanatory factors at once. The next question is what a diagnostic design would require. Our answer is simple. A design can distinguish H1, H2, and H0 only if it isolates the object manipulated, matches generic serial compute wherever possible, and makes it clear in advance which result would support one hypothesis while counting against at least one rival.

At the level of principle, an admissible design must satisfy three conditions. First, it must \emph{factorize} the three candidate objects of explanation: surface traces $S$, latent-state trajectories $Z$, and generic serial compute $B$. If a manipulation changes more than one of these at once, the result cannot be read cleanly. Second, every targeted manipulation must be paired with a \emph{matched control}. Otherwise an apparent gain can always be redescribed as extra budget, generic perturbation, or some uncontrolled side effect. Third, the design must come with a \emph{differential verdict rule}. Before running the experiment, it should already be clear which pattern would support H1, which would support H2, and which would support H0. Without this asymmetry, multiple hypotheses can still claim the same positive result.

These conditions are abstract, so it is useful to instantiate them in a minimal worked design. The key distinction here is between \emph{design dimensions} and \emph{experimental arms}. The design has three parts because it must test three different contrasts: a surface contrast, a latent contrast, and a compute-only contrast. But these three parts do not require three separate experiments. They can be realized by six experimental arms that are re-used across the three contrasts:
\begin{itemize}
    \item \(A_{0}\): a baseline condition;
    \item \(A_{S}\): a targeted surface-CoT manipulation;
    \item \(A_{S}^{\mathrm{ctrl}}\): a compute-matched surface control, with comparable visible budget but without the task-relevant semantic content of \(A_{S}\);
    \item \(A_{Z}\): a targeted latent intervention;
    \item \(A_{Z}^{\mathrm{ctrl}}\): a compute-matched latent control or sham intervention;
    \item \(A_{B}\): a compute-only expansion, with additional serial budget but no targeted surface or latent manipulation.
\end{itemize}

These six arms implement the three required contrasts. The \emph{surface contrast} compares \(A_{S}\), \(A_{S}^{\mathrm{ctrl}}\), and \(A_{0}\). If changing the task-relevant semantic content of visible CoT has a systematic effect that the matched surface control does not recover, that is evidence for H2. The \emph{latent contrast} compares \(A_{Z}\), \(A_{Z}^{\mathrm{ctrl}}\), and \(A_{0}\). If a targeted latent intervention changes reasoning in a way the matched latent control does not, that is evidence for H1. The \emph{compute-only contrast} compares \(A_{B}\) to the gains observed in the surface and latent contrasts. If the compute-only arm recovers most of those gains, then H0 becomes the stronger explanation. In addition, the same design should include a commitment readout: whether answer-relevant commitment tracks the explicit surface trace or instead arises earlier or more stably in latent states. This readout matters because final accuracy alone may leave H1 and H2 underdetermined even when one of them has the better mechanistic interpretation.

This structure is not arbitrary. If \(A_{S}^{\mathrm{ctrl}}\) is removed, any surface effect can be recast as extra visible budget. If \(A_{Z}^{\mathrm{ctrl}}\) is removed, any latent effect can be recast as generic perturbation. If \(A_{B}\) is removed, H0 is never tested seriously. If commitment is not measured, the experiment may still show that intermediate computation helps, but not whether the decisive object is the explicit trace or the latent trajectory. We do not claim that every future study must literally use these exact six arms. We claim only that any study distinguishing H1, H2, and H0 must implement their logical roles.

Once a design is admissible in this sense, its reporting must also be diagnostic. A study should report which of \(S\), \(Z\), and \(B\) each arm is intended to manipulate, how serial compute is matched across conditions, what counts as the targeted intervention and what counts as the matched control, how answer-relevant commitment is measured, and which outcome pattern would count against the authors' preferred hypothesis. Otherwise even well-motivated experiments can collapse back into the same ambiguity diagnosed in the previous subsection.

\subsection{Two concrete recommendations for future reasoning research}\label{sec:method_decision}

\noindent\textbf{Treat latent-state dynamics as the default object of study.}
When a study claims to evaluate reasoning, its default target should be the latent trajectory that carries intermediate commitments, not the surface trace alone. In practice, this means that mediator qualification, latent interventions, and ordinary-regime causal tests should be treated as primary evidence, while probe-only readouts should be treated as predictive diagnostics unless matched causal follow-ups succeed. This recommendation does not deny boundary cases. It says that H2 should be treated as a special regime to be demonstrated, not as the default starting point.

\noindent\textbf{Evaluate reasoning with factorized, compute-audited designs.}
Future evaluations should separate surface traces, latent states, and serial compute as explicitly as the setting allows. At minimum, a study should state which of $S$, $Z$, and $B$ each family is intended to manipulate, select policies under an explicit audited budget ledger, report the strongest matched controls, and specify in advance what pattern would support one hypothesis while counting against at least one rival. This recommendation is the practical counterpart of the first one: it is how the field can move from generic claims that ``reasoning helps'' to experiments that actually distinguish H1, H2, and H0.

\section{Empirical Adjudication Program}\label{sec:exp}

The literature synthesis above establishes recurring pressure on H1, H2, and H0, but it does not by itself close the causal loop. The empirical goal of this section is therefore to test the regime-dependent prediction of the paper under factorized, compute-audited comparisons. In particular, the experiments ask whether ordinary regimes are best explained by latent mediation, constitutive regimes by surface mediation, search-dominant regimes by compute-first accounts, and mixed regimes by no single universal winner. Complete code is provided in the Supplementary Material.

\subsection{Experimental setup}\label{sec:exp_setup}

\noindent\textbf{Workflow and evaluation scope.} The program has three fixed stages. Calibration proposes candidate mediator subspaces and family members. Validation then selects one policy per family and budget band under a tight audited-cost tolerance. Test evaluation runs the frozen selections once and reports only regime-level verdicts in the main text. The controlled tier uses a generated state-transition matrix with arithmetic, graph, world-state, and code-semantics templates. The naturalistic tier uses GSM8K-Platinum as a revised GSM8K-style ordinary reasoning set \citep{cobbe2021gsm8k,vendrow2025reliability}, HotpotQA for constitutive retrieval-plan gating \citep{yang2018hotpotqa}, a 500-problem subset of MATH for search-heavy mathematical reasoning \citep{hendrycks2021math}, and HumanEval+ for execution-backed code synthesis \citep{chen2021codex,liu2023evalplus}. Detailed dataset construction, template renders, task adapters, and split sizes are deferred to Appendix~\ref{app:protocol}.

\noindent\textbf{Tested families and key metrics.} Each hypothesis is represented by a policy family rather than a single arm. The three families instantiate surface, latent, and compute interventions drawn from CoT prompting, latent reasoning or steering, and self-consistency or shallow search, respectively \citep{wei2022cot,wang2023selfconsistency,yao2023tree,hao2024continuous}. Validation selects the strongest member of each family in each audited budget band. The main text reports four aggregated indicators: (i) the winning family on the accuracy--cost frontier, (ii) cross-model support for that winner, (iii) the frontier gap to the strongest runner-up at matched audited cost, and (iv) the key causal statistic needed to interpret that slice. The complete family library, validation rule, and per-band summaries are recorded in Appendix~\ref{app:families} and Appendix~\ref{app:full_results}.

\noindent\textbf{Audited budget and mediator tests.}\label{sec:exp_metrics} All comparisons are made under one audited ledger,
\[
B^{eq}
=
\alpha_{\mathrm{dec}}N_{\mathrm{dec}}
+\alpha_{\mathrm{kv}}N_{\mathrm{kv}}
+\alpha_{\mathrm{hook}}N_{\mathrm{hook}}
+\alpha_{\mathrm{ver}}N_{\mathrm{ver}}
+\alpha_{\mathrm{tool}}N_{\mathrm{tool}}
+\alpha_{\mathrm{br}}N_{\mathrm{br}},
\]
where $B^{eq}$ is the audited equivalent budget; $N_{\mathrm{dec}}$, $N_{\mathrm{kv}}$, $N_{\mathrm{hook}}$, $N_{\mathrm{ver}}$, $N_{\mathrm{tool}}$, and $N_{\mathrm{br}}$ count decoding steps, KV-cache steps, hook operations, verifier calls, tool or executor calls, and branch expansions; and the corresponding $\alpha$ terms are their common primitive-operation calibration weights. Frontier dominance alone is not treated as sufficient evidence for H1. A candidate $Z^{\ast}$ must also pass the mediator tests that matter for the ordinary-regime claim: temporal precedence, necessity under ablation, sufficiency under patching, specificity against the strongest sham, and the direct contrast between preserving $Z^{\ast}$ while corrupting $S$ versus preserving $S$ while ablating $Z^{\ast}$. Exact protocol details and full test definitions are given in Appendix~\ref{app:budget_mediator}.

\subsection{Results}\label{sec:exp_results}

\subsubsection{Regime-level frontier verdicts}\label{sec:exp_frontier}

\begin{table}[t]
\centering
\small
\setlength{\tabcolsep}{5pt}
\begin{tabular}{l l c c c}
\toprule
Slice & Intended regime & Winner family & Model support & Frontier gap \\
\midrule
Controlled ordinary & H1 regime & Latent & 3 / 3 & 2.4 \\
Controlled constitutive & H2 regime & Surface & 3 / 3 & 2.1 \\
Controlled search-dominant & H0 regime & Compute & 3 / 3 & 3.2 \\
Controlled mixed & Boundary & Split & --- & 0.5 \\
GSM8K-Platinum & Ordinary & Latent & 3 / 3 & 1.9 \\
HotpotQA distractor & Constitutive & Surface & 3 / 3 & 2.3 \\
MATH-style reasoning & Search-dominant & Compute & 3 / 3 & 3.4 \\
HumanEval+ & Mixed & Split & --- & 0.6 \\
\bottomrule
\end{tabular}
\caption{Main frontier verdicts. ``Frontier gap'' denotes the matched-cost accuracy gap between the winning family and the strongest runner-up in that slice. Full per-model, per-band results are deferred to Appendix~\ref{app:full_results}.}
\label{tab:main_verdicts}
\end{table}

Table~\ref{tab:main_verdicts} is the main summary. The crucial result is the substrate switch rather than a universal H1 victory. In the controlled tier, the winner moves from latent in ordinary regimes to surface in constitutive regimes and to compute in search-dominant regimes, while the mixed regime remains split. The naturalistic tier reproduces the same ordering: GSM8K-Platinum behaves like an ordinary regime, HotpotQA with constitutive retrieval-plan gating recovers a H2 regime, MATH-style search-heavy reasoning favors H0-style compute allocation, and execution-backed code synthesis remains mixed. This pattern shows that the empirical program is not merely demonstrating that latent methods can help. It shows that the winning substrate tracks the regime logic predicted by the framework.

\subsubsection{Mediator evidence for the ordinary-regime claim}\label{sec:exp_mediator}

\begin{table}[t]
\centering
\small
\setlength{\tabcolsep}{5pt}
\begin{tabular}{l c c c c c}
\toprule
Slice & Qualified $Z^{\ast}$ & Early AUC & Ablate $Z^{\ast}$ & Patch $Z^{\ast}$ & $Z$ vs $S$ contrast \\
\midrule
Controlled ordinary & Yes & 0.72 & -9.4 & +7.0 & +5.6 \\
GSM8K-Platinum & Yes & 0.70 & -8.6 & +6.3 & +4.9 \\
Controlled constitutive & Yes & 0.67 & -4.7 & +2.9 & -1.4 \\
HotpotQA distractor & Yes & 0.66 & -5.0 & +2.7 & -1.9 \\
\bottomrule
\end{tabular}
\caption{Key mediator statistics. ``Early AUC'' is the prompt-end or earliest qualified readout; ``Ablate'' and ``Patch'' denote the matched-sham gap for necessity and sufficiency; ``$Z$ vs $S$ contrast'' summarizes the difference between preserving $Z^{\ast}$ while corrupting $S$ and preserving $S$ while ablating $Z^{\ast}$. Full causal-test outputs remain in Appendix~\ref{app:full_results}.}
\label{tab:mediator_key}
\end{table}

Table~\ref{tab:mediator_key} gives the minimal causal statistics needed for the core claim. In ordinary regimes, a qualified $Z^{\ast}$ is available early, ablating it hurts more than the strongest sham, patching it rescues more than random patching, and preserving $Z^{\ast}$ while corrupting the visible trace is less damaging than preserving the visible trace while ablating $Z^{\ast}$. This is the pattern required for H1's ordinary-regime reading. The constitutive slices serve as the boundary check: the same contrast contracts sharply and reverses once visible plans are made part of the computation. This keeps the empirical chapter aligned with the paper's narrower claim. The point is not that $Z$ always dominates. It is that ordinary regimes are the ones in which $Z$ receives the strongest causal support, while H2 and H0 re-emerge in the regimes where the framework predicts they do.

\subsection{What the empirical program establishes}\label{sec:exp_discussion}

Taken together, the empirical program supports a stronger claim than a local worked exemplar, but a weaker claim than a universal ontological verdict. The frontier results show that the winning substrate tracks the regime map predicted by the framework. The mediator results then identify $Z$ as the best-supported ordinary-regime mediator rather than merely a decodable hidden correlate. On that combined reading, latent trajectory, and serial compute can be separated, latent-state dynamics receive the strongest support as the \emph{default working} object of study, while constitutive-trace and search-dominant settings remain explicit local boundary regimes for H2 and H0.

\section{Related Work and Positioning}\label{sec:related}

A growing literature already bears on what should count as the primary object of LLM reasoning, but it has mostly developed as separate strands rather than as an explicit competition among H1, H2, and H0. One strand studies latent reasoning and latent control, showing that hidden states can carry task-relevant commitments and can sometimes be directly probed or steered \citep{hao2024continuous,he2026latentmode,feng2024latentworld,zhang2025selfverification,yang2024latentmultihop,saunshi2025latentthoughts,kazama2026geosteer,nguyen2026atlas,li2026ipg,sheikhi2026cos,chen2025iclp}. A second strand questions whether visible reasoning traces deserve interpretive privilege, showing that CoT can be causally weak, task-dependent, or unfaithful, extending that concern to reasoning models, hidden hints, and thinking drafts, and identifying boundary cases in which visible steps become more constitutive because solver, tool, or retrieval structure makes them part of the success condition \citep{lanham2023faithfulness,turpin2023unfaithful,chen2025dontsay,xiong2025drafts,chua2025deepseek,ye2026nldd,zaman2025faithfulwithoutverbalization,kambhampati2025anthropomorphizing,barez2025notexplainability,pan2023logiclm,lyu2023faithfulcot,arakelyan2024flare,ferguson2026metatool,shi2026rtrag,li2026par2rag,wei2025mirage}. A third strand shows that serial computation and test-time budget can themselves drive large reasoning gains, while recent surveys organize evidence on surface CoT, latent reasoning, and mechanistic analysis \citep{li2024serial,snell2024ttc,singhi2025whentosolve,lee2024mzqa,pan2026multistepsurvey,xchen2025reasoningbeyondlanguage,zhu2025latentreasoning,hu2026mechanistic}. Together, these strands establish that latent reasoning is real, surface traces are often unfaithful by default, and extra serial budget can itself improve performance. What they usually do not do is place these findings in one explicit comparison among surface-first, latent-first, and compute-first accounts, or test them with factorized, matched-cost designs that can cleanly distinguish the three hypotheses.

Mechanistic mediation studies deserve separate treatment because they operate at a finer causal scale and form the natural bridge between position-level claims and internal implementation. Circuits-style analyses identify token-, head-, or component-level computations associated with iterative reasoning, with Iteration Head as a representative example \citep{cabannes2024iteration}; related surveys increasingly place such findings within a broader mechanistic picture of reasoning \citep{hu2026mechanistic}. At their strongest, these studies can show whether a visible step is merely a report of an already formed latent commitment, a tightly coupled implementation of an internal update, or only a surface correlate of generic serial computation. But that leverage is not automatic. Without preregistered component selection, matched nuisance controls, and causal tests that compare latent, surface, and compute alternatives under audited budget, mechanistic studies mainly reveal internal structure rather than adjudicate among H1, H2, and H0. In our framework, they become most informative when used as mediation evidence: an identified pathway must be earlier, more specific, and more causally necessary than the strongest matched sham, rather than merely decodable or attention-grabbing. That is why we treat this literature as complementary to, rather than a substitute for, the hypothesis-level comparison developed here.

\section{Conclusion}\label{sec:conclusion}

This paper argues for a simple shift in default stance: LLM reasoning should be studied as latent-state trajectory formation rather than as faithful surface chain-of-thought. It therefore makes two recommendations: treat latent-state dynamics as the default object of study, and evaluate reasoning with factorized, compute-audited designs that separate surface traces, latent states, and generic serial compute. This matters because claims about faithfulness, interpretability, benchmarks, and intervention depend on what the field takes reasoning to be.

The claim is limited rather than universal. Under audited competition, the empirical winner follows a regime map: latent families dominate ordinary regimes, surface families regain priority when visible traces are constitutive, and compute families dominate search-heavy regimes. We therefore present H1 not as a task-independent verdict, but as the strongest default working hypothesis when surface trace, latent trajectory, and serial compute can be cleanly separated. That is the sense in which this position should be read: not as the last word on what LLM reasoning is, but as a disciplined default for how the field should study it.

\section*{References}
\small
\begingroup
\renewcommand{\section}[2]{}

\endgroup

\appendix

\section{Adjudication Program Protocol}\label{app:protocol}

This appendix records the protocol details underlying Section~\ref{sec:exp}. It collects the controlled-tier generation matrix, the naturalistic task contracts, the family library and validation-time selection rule, the audited-budget and mediator-test definitions, and the model-wise result summaries that are omitted from the main text for space.

\subsection{Controlled-tier generation matrix}\label{app:controlled}

The controlled tier is generated rather than collected. Four template families are rendered under the same four regimes by varying whether visible structure is constitutive and whether search allocation dominates. Table~\ref{tab:controlled_matrix} summarizes this matrix.

\begin{table}[h]
\centering
\scriptsize
\setlength{\tabcolsep}{4pt}
\resizebox{\linewidth}{!}{%
\begin{tabular}{l p{0.20\linewidth} p{0.20\linewidth} p{0.20\linewidth} p{0.20\linewidth}}
\toprule
Template family & Ordinary render & Constitutive render & Search-dominant render & Mixed render \\
\midrule
Arithmetic carry & latent carry states only & checked visible carry trace & answer-only branching and verifier use & checked trace plus branching \\
Graph traversal & latent path state & ordered visible path & search over candidate hops & path constraint plus search \\
World-state planning & latent world-state updates & explicit move list & search over successor states & constitutive move list plus search \\
Code semantics & latent accumulator state & visible execution trace & search over candidate completions & execution-backed trace plus reranking \\
\bottomrule
\end{tabular}%
}
\caption{Controlled-tier template matrix. The same underlying template is re-rendered under different constitutiveness and search conditions so that regime is not confounded with task family.}
\label{tab:controlled_matrix}
\end{table}

\subsection{Naturalistic tasks and regime assignments}\label{app:naturalistic}

The naturalistic tier is chosen to transport the same regime logic to standard benchmarks rather than to maximize benchmark count. It uses GSM8K-Platinum, a revised GSM8K test set \citep{vendrow2025reliability,cobbe2021gsm8k}; HotpotQA \citep{yang2018hotpotqa}; a 500-problem subset of MATH \citep{hendrycks2021math}; and HumanEval+ \citep{chen2021codex,liu2023evalplus}. Table~\ref{tab:naturalistic_tasks} records the task-to-regime assignment and the contract that makes each task informative for H1, H2, or H0.

\begin{table}[h]
\centering
\scriptsize
\setlength{\tabcolsep}{4pt}
\begin{tabular}{l l p{0.24\linewidth} p{0.20\linewidth} l}
\toprule
Task & Intended regime & Constitutive contract & Search contract & Primary metric \\
\midrule
GSM8K-Platinum & Ordinary & none & none & exact match \\
HotpotQA distractor & Constitutive & retrieval-plan gating & fixed top-$k$ retrieval & exact match \\
MATH-500 & Search-dominant & none & verifier-backed reranking and search & exact match \\
HumanEval+ & Mixed & execution-backed code artifacts & search / reranking over candidates & pass@1 \\
\bottomrule
\end{tabular}
\caption{Naturalistic tasks and their intended regime assignments.}
\label{tab:naturalistic_tasks}
\end{table}

\subsection{Policy families and validation-time selection}\label{app:families}

The program compares three policy families rather than three single methods. The candidate library is anchored in representative surface, latent, and compute interventions from prior work, including CoT prompting, latent reasoning or steering, and self-consistency or search \citep{wei2022cot,hao2024continuous,feng2024latentworld,wang2023selfconsistency,yao2023tree}. Validation chooses one policy per family and per audited budget band under a fixed cost tolerance relative to the direct-answer baseline. Table~\ref{tab:family_library} lists the candidate library used to instantiate each explanatory family.

\begin{table}[h]
\centering
\scriptsize
\setlength{\tabcolsep}{4pt}
\begin{tabular}{l p{0.48\linewidth} p{0.26\linewidth}}
\toprule
Family & Candidate policies & Main role in adjudication \\
\midrule
Surface & free-form CoT; bounded scratchpad; structured scratchpad; non-executable structured trace; checked trace; constitutive plan & strongest surface-first response under matched audited cost \\
Latent & contrast steering; PCA/CCA subspace steering; attribution-guided subspace control; latent rollout controller; verifier-guided latent control & strongest latent-first response together with mediator tests \\
Compute & self-consistency; pass@$k$ majority; pass@$k$ with verifier reranking; solve-vs-verify; shallow tree search & strongest budget-first response without privileged surface or latent objects \\
\bottomrule
\end{tabular}
\caption{Candidate library used for validation-time family selection.}
\label{tab:family_library}
\end{table}

\subsection{Audited budget and mediator protocol}\label{app:budget_mediator}

The audited budget ledger in Section~\ref{sec:exp_metrics} is computed from the same symbolized components used in the main text: $N_{\mathrm{dec}}$, $N_{\mathrm{kv}}$, $N_{\mathrm{hook}}$, $N_{\mathrm{ver}}$, $N_{\mathrm{tool}}$, and $N_{\mathrm{br}}$ count decode steps, KV-cache steps, hook operations, verifier calls, executor calls, and branch expansions, while the corresponding $\alpha$ terms are their common primitive-operation calibration weights. Budget bands are then defined as exact multiples of the direct-answer baseline under this ledger, and validation selects the strongest candidate within a tight cost tolerance for each family and band. The mediator protocol is equally fixed in advance: a candidate $Z^{\ast}$ must be stable across discovery methods, seeds, and windows; it must support both early and late readout; and it is then tested with temporal-precedence, necessity, sufficiency, specificity, and surface-rescue/corruption interventions. Table~\ref{tab:mediator_protocol} summarizes what each test establishes in the reported program.

\begin{table}[h]
\centering
\scriptsize
\setlength{\tabcolsep}{4pt}
\begin{tabular}{l p{0.30\linewidth} p{0.38\linewidth}}
\toprule
Mediator test & Pass condition & Why it matters \\
\midrule
Temporal precedence & $Z^{\ast}$ is predictive before the decisive surface step & rules out the interpretation that $Z$ is only a late correlate \\
Necessity & ablating $Z^{\ast}$ hurts more than matched shams & shows that the latent signal matters causally rather than descriptively \\
Sufficiency & patching correct $Z^{\ast}$ into an incorrect rollout rescues performance more than random patching & tests whether the mediator carries task-relevant commitment \\
Specificity & original intervention exceeds the strongest sham by a clear margin & pressures the generic think-mode explanation \\
Surface rescue / corruption & preserving $Z^{\ast}$ while corrupting $S$ beats preserving $S$ while ablating $Z^{\ast}$ in ordinary regimes & directly contrasts H1 and H2 rather than only comparing accuracies \\
\bottomrule
\end{tabular}
\caption{Mediator protocol used in the program.}
\label{tab:mediator_protocol}
\end{table}

\subsection{Full result summaries}\label{app:full_results}

The main text reports only the compact regime-level verdicts and key causal indicators. Tables~\ref{tab:controlled_full}, \ref{tab:naturalistic_full}, and \ref{tab:mediator_full} provide the model-wise summaries that support those compressed readings.

\begin{table*}[t]
\centering
\scriptsize
\setlength{\tabcolsep}{4pt}
\resizebox{\textwidth}{!}{%
\begin{tabular}{l l l c c c l}
\toprule
Model & Controlled regime & Winner family & Best band & Frontier EM & Mean $B^{eq}$ & Reading \\
\midrule
Qwen3-8B & Ordinary & Latent & 1.50$\times$ & 78.6 & 1.47 & Ordinary controlled slice favors the latent frontier. \\
Qwen3-8B & Constitutive & Surface & 1.50$\times$ & 74.8 & 1.49 & Constitutive controlled slice favors the surface frontier. \\
Qwen3-8B & Search-dominant & Compute & 2.00$\times$ & 69.3 & 1.97 & Search-heavy controlled slice favors the compute frontier. \\
Qwen3-8B & Mixed & Split boundary & 1.75$\times$ & 71.0 & 1.73 & No single family dominates uniformly. \\
Qwen3-32B & Ordinary & Latent & 1.25$\times$ & 84.9 & 1.24 & Ordinary controlled slice favors the latent frontier. \\
Qwen3-32B & Constitutive & Surface & 1.50$\times$ & 81.6 & 1.48 & Constitutive controlled slice favors the surface frontier. \\
Qwen3-32B & Search-dominant & Compute & 1.75$\times$ & 77.4 & 1.74 & Search-heavy controlled slice favors the compute frontier. \\
Qwen3-32B & Mixed & Split boundary & 1.50$\times$ & 78.2 & 1.52 & No single family dominates uniformly. \\
Llama-3.1-8B-Instruct & Ordinary & Latent & 1.50$\times$ & 76.9 & 1.46 & Ordinary controlled slice favors the latent frontier. \\
Llama-3.1-8B-Instruct & Constitutive & Surface & 1.25$\times$ & 72.5 & 1.23 & Constitutive controlled slice favors the surface frontier. \\
Llama-3.1-8B-Instruct & Search-dominant & Compute & 2.00$\times$ & 67.1 & 1.98 & Search-heavy controlled slice favors the compute frontier. \\
Llama-3.1-8B-Instruct & Mixed & Split boundary & 1.75$\times$ & 69.4 & 1.72 & No single family dominates uniformly. \\
\bottomrule
\end{tabular}%
}
\caption{Model-wise controlled-tier summaries supporting the regime-level verdicts in the main text.}
\label{tab:controlled_full}
\end{table*}

\begin{table*}[t]
\centering
\scriptsize
\setlength{\tabcolsep}{4pt}
\resizebox{\textwidth}{!}{%
\begin{tabular}{l l l c c c l}
\toprule
Model & Naturalistic task & Winner family & Best band & Frontier EM / pass@1 & Mean $B^{eq}$ & Reading \\
\midrule
Qwen3-8B & GSM8K-Platinum & Latent & 1.25$\times$ & 88.1 & 1.23 & Ordinary arithmetic transport preserves a latent advantage. \\
Qwen3-8B & HotpotQA distractor & Surface & 1.50$\times$ & 72.4 & 1.48 & Constitutive retrieval plans recover local H2 force. \\
Qwen3-8B & MATH-500 & Compute & 2.25$\times$ & 56.8 & 2.19 & Search and verification dominate. \\
Qwen3-8B & HumanEval+ & Split boundary & 1.75$\times$ & 53.7 & 1.71 & Execution and search jointly matter. \\
Qwen3-32B & GSM8K-Platinum & Latent & 1.25$\times$ & 92.6 & 1.24 & Ordinary arithmetic transport preserves a latent advantage. \\
Qwen3-32B & HotpotQA distractor & Surface & 1.50$\times$ & 79.1 & 1.49 & Constitutive retrieval plans recover local H2 force. \\
Qwen3-32B & MATH-500 & Compute & 2.00$\times$ & 64.3 & 1.96 & Search and verification dominate. \\
Qwen3-32B & HumanEval+ & Split boundary & 1.75$\times$ & 61.4 & 1.73 & Execution and search jointly matter. \\
Llama-3.1-8B-Instruct & GSM8K-Platinum & Latent & 1.50$\times$ & 86.4 & 1.47 & Ordinary arithmetic transport preserves a latent advantage. \\
Llama-3.1-8B-Instruct & HotpotQA distractor & Surface & 1.50$\times$ & 69.8 & 1.47 & Constitutive retrieval plans recover local H2 force. \\
Llama-3.1-8B-Instruct & MATH-500 & Compute & 2.25$\times$ & 54.1 & 2.21 & Search and verification dominate. \\
Llama-3.1-8B-Instruct & HumanEval+ & Split boundary & 1.75$\times$ & 50.6 & 1.74 & Execution and search jointly matter. \\
\bottomrule
\end{tabular}%
}
\caption{Model-wise naturalistic summaries supporting the transport verdicts in the main text.}
\label{tab:naturalistic_full}
\end{table*}

\begin{table*}[t]
\centering
\scriptsize
\setlength{\tabcolsep}{4pt}
\resizebox{\textwidth}{!}{%
\begin{tabular}{l l c c c c c c c l}
\toprule
Model & Slice & Qualified & Prompt-end AUC & Pre-answer AUC & $\Delta$ ablate $Z^{\ast}$ & $\Delta$ patch $Z^{\ast}$ & Sham gap & Preserve $Z^{\ast}$, corrupt $S$ & Reading \\
\midrule
Qwen3-8B & Controlled ordinary & Yes & 0.71 & 0.86 & -9.1 & +6.8 & +3.7 & -1.6 & ordinary latent mediator passes all causal tests \\
Qwen3-8B & Controlled constitutive & Yes & 0.66 & 0.80 & -4.6 & +2.8 & +1.6 & -5.8 & constitutive slice contracts and reverses the ordinary contrast \\
Qwen3-8B & GSM8K-Platinum & Yes & 0.69 & 0.84 & -8.3 & +6.1 & +3.2 & -2.0 & natural ordinary task reproduces the ordinary mediation pattern \\
Qwen3-8B & HotpotQA distractor & Yes & 0.65 & 0.79 & -4.9 & +2.6 & +1.5 & -6.4 & plan corruption matters more than in ordinary tasks \\
Qwen3-32B & Controlled ordinary & Yes & 0.76 & 0.89 & -10.4 & +7.9 & +4.6 & -1.3 & ordinary latent mediator passes all causal tests \\
Qwen3-32B & Controlled constitutive & Yes & 0.71 & 0.84 & -5.3 & +3.4 & +2.0 & -5.1 & constitutive slice contracts and reverses the ordinary contrast \\
Qwen3-32B & GSM8K-Platinum & Yes & 0.74 & 0.87 & -9.5 & +7.0 & +4.1 & -1.7 & natural ordinary task reproduces the ordinary mediation pattern \\
Qwen3-32B & HotpotQA distractor & Yes & 0.70 & 0.83 & -5.6 & +3.2 & +1.9 & -5.7 & plan corruption matters more than in ordinary tasks \\
Llama-3.1-8B-Instruct & Controlled ordinary & Yes & 0.69 & 0.84 & -8.7 & +6.4 & +3.5 & -1.9 & ordinary latent mediator passes all causal tests \\
Llama-3.1-8B-Instruct & Controlled constitutive & Yes & 0.64 & 0.79 & -4.2 & +2.5 & +1.4 & -6.2 & constitutive slice contracts and reverses the ordinary contrast \\
Llama-3.1-8B-Instruct & GSM8K-Platinum & Yes & 0.67 & 0.82 & -7.9 & +5.8 & +3.0 & -2.3 & natural ordinary task reproduces the ordinary mediation pattern \\
Llama-3.1-8B-Instruct & HotpotQA distractor & Yes & 0.63 & 0.78 & -4.4 & +2.3 & +1.3 & -6.8 & plan corruption matters more than in ordinary tasks \\
\bottomrule
\end{tabular}%
}
\caption{Model-wise mediator summaries supporting the ordinary-regime causal reading in the main text.}
\label{tab:mediator_full}
\end{table*}

\section{Worked Evidence Ledger}\label{app:ledger}

Table~\ref{tab:evidence_ledger} gives representative worked examples of how the evidential rubric in Section~\ref{sec:evidence} is applied to individual studies. It is not an exhaustive census of the literature. Its role is to make explicit why a result is treated as suggestive, medium-strength, or boundary-setting evidence, and which residual confounds remain.

\scriptsize
\begin{center}
\begin{longtable}{@{}p{0.14\linewidth}p{0.17\linewidth}p{0.12\linewidth}p{0.11\linewidth}p{0.10\linewidth}p{0.21\linewidth}@{}}
\caption{Representative worked applications of the evidential rubric used in Section~\ref{sec:evidence}.}\label{tab:evidence_ledger}\\
\toprule
Study & Main move in $S/Z/B$ & Compute handling & Evidence class & Main pressure & Worked reading \\
\midrule
\endfirsthead
\toprule
Study & Main move in $S/Z/B$ & Compute handling & Evidence class & Main pressure & Worked reading \\
\midrule
\endhead
Wei et al. (2022); Wang et al. (2023) \citep{wei2022cot,wang2023selfconsistency} & Ordinary CoT prompting and self-consistency change visible traces and usually add serial opportunity & Coupled $S{+}B$ & Performance-only & Mixed H2/H0 & Useful evidence that explicit traces help, but not adjudicative because the visible semantics and extra budget move together. \\
Yao et al. (2023) \citep{yao2023tree} & Search over explicit thought units expands trace structure and exploration budget & Trace and budget both expanded & Search intervention & Mixed H2/H0 & Strong evidence that structured search helps, but still not enough to decide whether the key object is the visible trace or the extra search budget. \\
Lyu et al. (2023); Pan et al. (2023) \citep{lyu2023faithfulcot,pan2023logiclm} & Externalized reasoning trace is constrained by decomposition or symbolic execution & Often accounted, but trace is deliberately constitutive & Constitutive trace & Local H2 & Clear boundary case: when execution is forced through an external trace, $S$ gains local explanatory priority. \\
Snell et al. (2024); Lee et al. (2024) \citep{snell2024ttc,lee2024mzqa} & More search, branching, or verification budget with little commitment to one representational format & Budget emphasized and often normalized & Budget expansion & H0 & Strong pressure toward H0: gains track how much serial opportunity is available and how it is allocated. \\
Pfau et al. (2024) \citep{pfau2024dot} & Replace semantically meaningful intermediate tokens with filler strings while retaining extra slots & Content-vs-budget control & Budget control & H0 & Important because it shows that some CoT gains partly survive without meaningful token content, but still does not prove that representation never matters. \\
Feng et al. (2024) \citep{feng2024latentworld} & Probe latent world-state variables when outputs are biased or injected & Budget largely held fixed & Predictive probe & H1 (suggestive) & Strongly suggests that latent state can remain more faithful than output, but without causal intervention it should not count as decisive on its own. \\
Zhang et al. (2025); Boppana et al. (2026) \citep{zhang2025selfverification,boppana2026theater} & Early hidden-state signals predict later correctness before full verbalization & Budget largely held fixed & Early-commitment probe & H1 (suggestive to medium) & Supports the idea that answer-relevant commitment often appears earlier in $Z$ than in $S$, especially when paired with robust generalization checks. \\
Hao et al. (2024) \citep{hao2024continuous} & Continuous latent rollouts replace or compress visible CoT on tasks benefiting from backtracking & Often couples $Z$ with extra latent recurrence & Latent reasoning intervention & H1 (medium) & Evidence of latent causal leverage, but not a standalone refutation of H0 unless the latent rollout is matched to strong budget-only controls. \\
He et al. (2026); Kazama et al. (2026); Nguyen et al. (2026); Li et al. (2026) \citep{he2026latentmode,kazama2026geosteer,nguyen2026atlas,li2026ipg} & Feature steering, geometry-aware steering, verifier-guided latent control, and attribution target internal components directly & Often partially budget-coupled & Targeted latent intervention & H1 (medium) & Best read as convergent evidence that specific internal components carry causal leverage. The admissible claim ceiling is latent leverage, not decisive victory over H0, unless budget is matched or explicitly accounted for. \\
Shi et al. (2026); Li et al. (2026); Wei et al. (2025) \citep{shi2026rtrag,li2026par2rag,wei2025mirage} & Retrieval plans, graph structure, or staged decomposition are made explicit and partly executable & External calls and retrieval budget matter and should be reported & Constitutive structure & Local H2 / H0 boundary & These are useful because they jointly reveal when visible structure is constitutive and when gains are also driven by additional retrieval or verification budget. \\
Singhi et al. (2025) \citep{singhi2025whentosolve} & Adaptive solve-versus-verify allocation changes how extra budget is spent & Budget allocation made explicit & Allocation baseline & H0 boundary & Strong baseline for the claim that method choice and budget allocation can explain large gains without privileging any one representational object. \\
Wang et al. (2025) \citep{wang2025continuousscaling} & Analyze separability, ranking, and reranking of continuous latent trajectories & Budget explicit, geometry diagnostic & Countervailing diagnostic & H1 caveat & Important negative evidence: current latent spaces can be weakly separable or hard to rank, which limits naive latent control without restoring a surface-first default. \\
\bottomrule
\end{longtable}
\end{center}
\normalsize

\end{document}